\documentclass{article}
\usepackage{ijcai23}

\usepackage{times}
\usepackage{soul}
\usepackage{url}
\usepackage[hidelinks]{hyperref}
\usepackage[utf8]{inputenc}
\usepackage[small]{caption}
\usepackage{graphicx}
\usepackage{amsmath}
\usepackage{amsthm}
\usepackage{booktabs}
\usepackage{algorithm}
\usepackage{algorithmic}
\usepackage[switch]{lineno}

\usepackage{amsfonts}
\usepackage{comment}
\usepackage{multirow}
\usepackage{bbm}
\usepackage{graphicx}
\usepackage{mathrsfs}
\usepackage{xcolor}

\title{Graph Relation Aware Continual Learning}

\author{
Qinghua Shen$^1$
\and
Weijieying Ren$^2$\and
Wei Qin$^{2,3}$
\affiliations
$^1$University of Science and Technology of China\\
$^2$The Pennsylvania State University\\
$^3$Hefei University of Technology\\
\emails
sqh2017ustc@mail.ustc.edu.cn,
wjr5337@psu.edu,
qinwei.hfut@gmail.com
}

\begin{document}

\maketitle
\begin{abstract}
Continual graph learning (CGL) studies the problem of learning from an infinite stream of graph data, 
consolidating historical knowledge,
and generalizing it to the future task. 
At once, only current graph data are available. 
Although some recent attempts have been made to handle this task, we still face two potential challenges: 
1) most of existing works only manipulate on the intermediate graph embedding and ignore intrinsic properties of graphs. It is non-trivial to differentiate the transferred information across graphs.
2) recent attempts take a parameter-sharing policy to transfer knowledge across time steps or progressively expand new architecture given shifted graph distribution. Learning a single model could loss discriminative information for each graph task while the model expansion scheme suffers from high model complexity.
In this paper, we point out that latent relations behind graph edges can be attributed as an invariant factor for the evolving graphs and the statistical information of latent relations evolves. 
Motivated by this, we design a relation-aware adaptive model, dubbed as RAM-CG, that consists of a relation-discovery modular to explore latent relations behind edges and a task-awareness masking classifier to accounts for the shifted. Extensive experiments show that
RAM-CG provides significant 2.2$\%$, 6.9 $\%$ and 6.6$\%$ accuracy improvements over the state-of-the-art results on 
CitationNet, OGBN-arxiv and TWITCH dataset, respective.

\end{abstract}
\section{Introduction}
Graph neural network \cite{DBLP:journals/corr/KipfW16,zhao2022exploring}, as a typical paradigm to represent graph-structured data, provides a universal framework to update each node representation via aggregating neighbors' information and the representation of the node itself \cite{simon20agg}. 
Significant methodological advances have been made in the field of GNN \cite{zhao2023faithful}, which have achieved promising performance in recommendation systems \cite{wang2022explanation}, traffic prediction \cite{traffic}, and drug design \cite{drug_design}, etc.

\begin{figure}[t]
\centering
\includegraphics[scale=0.35]{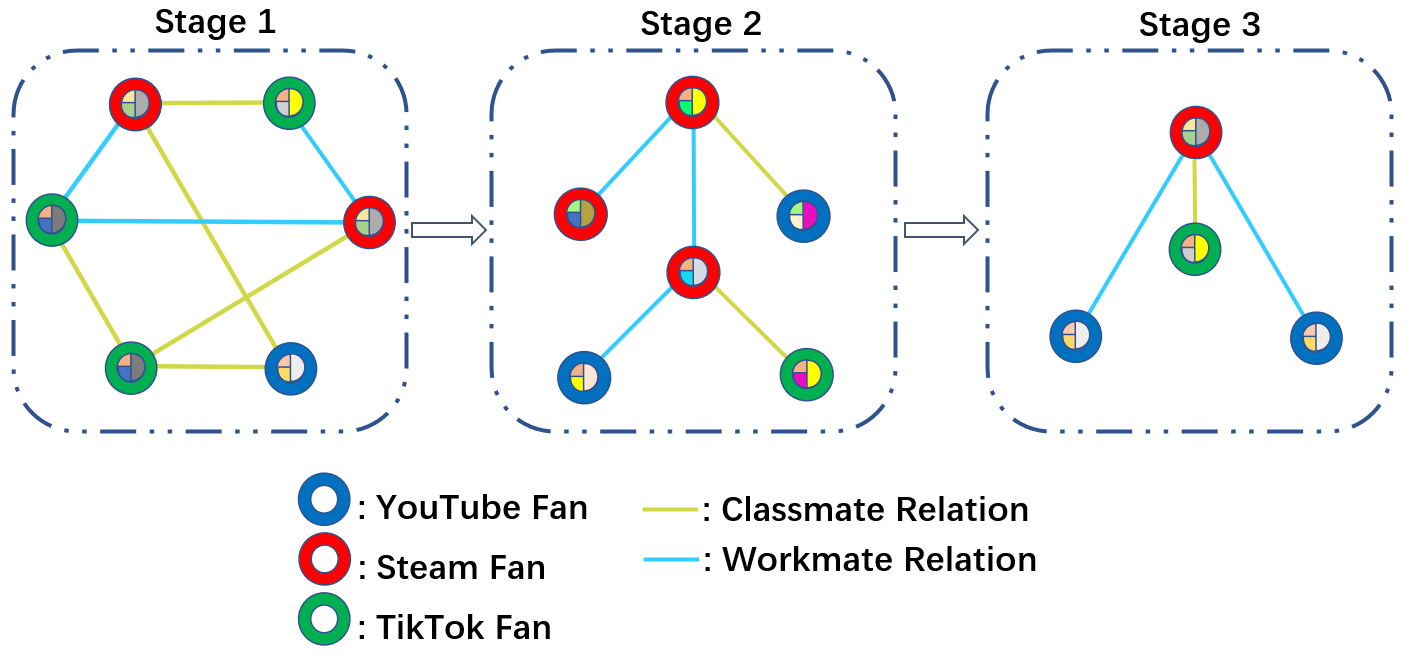}
\caption{An illustration of shifted social network. The diagram presents a  person's social network in three time stages. The outside ring of the graph node represents different preferences in different colours, while the inner sectors present the personal features. Under the out of distribution graph sequence scenario, not only the graph topology but also the node features shift apparently. However, the logic of building edge consistently focuses on the extension of Classmate Relation and Workmate Relation.}
\label{fig1}
\end{figure}

Recent works are mainly built on the I.I.D. hypothesis \cite{ren2017robust}, i.e., training and test graph data are independent and identically distributed. Existing literatures suggests that advanced GNN-based methods (e.g., GCN, GAT) lack the out-of-distribution (OOD) generalization ability \cite{gui2022good,ren2021fair,zhao2020semi}.
Taking the social network as a motivating example (as shown in Figure \ref{fig1}), {both the social relations and personal profiles of a specific user would change with time \cite{ren2022mitigating}. Besides the influence of aging, the rapid-developing society would also change the individual lifestyle.
Young people nowadays hardly preserve the same habits or values as their elder brothers and sisters used to.}
This example shows two distinct properties of evolving graph scenarios: 1) graph data come easily in a growing manner;
2) graph distribution shifts in real-world scenarios. Therefore, there is an ever-increasing demand for developing approaches capable of handling OOD generalization on graphs, specifically learn from a non-stationary stream of graph-related tasks without forgetting previous knowledge \cite{li2022learning}.

Motivated by the success of continual learning \cite{lopez2017gradient},
several recent continual graph learning (CGL) methods are proposed to effectively tackle knowledge transfer in graph-related data \cite{ego,graphnas,ExpandableGNN}.
Recent works are designed on one static model or progressive models such that is accurate (e.g., high classification accuracy on current data), memorable (e.g., easily retrieving historical graph patterns), and generalizable (e.g., coordinating and adapting on unseen graph data). Despite its effectiveness, most of these methods only manipulate on the intermediate graph embeddings and ignore the intrinstic property of graph data \cite{ego}, results in low effectiveness on newly coming shifted graph data. 
Besides, to further leverage distinctive information for each of shifted graph, recent work \cite{ExpandableGNN,graphnas} progressively expand new architecture when detecting new graph data and leveraging the reinforcement learning strategy to determine an optimal number of nodes to be added/pruned from the base network. 
However, these model expansion methods suffer from considerable network complexity and can not guarantee an optimal resource allocation. 

We backtrack two key points of the above two problems. 
First, what information is beneficial to be transferred in the evolving graphs? 
If we can 1) characterize what signals from the graph structure we can learn in the dynamic environment, and 2) differentiate the invariant information that is beneficial to be transferred across graph-related tasks and the shifted information that requires customized modular design, we can design a mode accurate model for the evolving graphs. 
Second, how to compose the two information in a seemingly manner, thus we can leverage the less model parameters that can generalize well on each new graph data?

To answer these questions, we propose a RAM-CG framework that is generalizable to newly incoming data over evolving graphs. 
We show that by discovering latent relations behind edge existence, dynamics over the evolving graphs can be well-encoded with a relation-based self-modulating graph neural network \cite{zhao2023skill}. Semantics of relations are \textbf{invariant} throughout time steps while their predictive power towards node labels would \textbf{vary} over time, which motivates our design of RAM-CG.
RAM-CG consists of two modules that separately process these two different kinds of information. 
1) relation discovery module, which disentangles latent relations behind graph edges; 2) task-awareness masking classifier, which learn an masking-based adaptive classifier in order to improve the effectiveness of the learned model. The insight here is that the statistical information of the latent relations may shift over time. To assign different weights to different evolving latent relation patterns, we train an adaptive classifier for each shifted graphs.

In particular, we make the following contributions:

\begin{itemize}
\item We propose the relation-aware adaptive model for continual graph learning, RAM-CG, a novel transfer learning method which can address three challenges: 1) graph comes in an sequence manner; 2) graph distribution shift across timelines; 3) each time, only current graph data are available at once. 
\item We find that latent relations behind edges can be attributed as invariant factors in the evolving graphs. Besides,  the statistical information of the latent
relations may shift over time and can be estimated by an adaptive classifier.
\item Extensive experiments on continual graph learning setting shows that our method is highly generalizable, accurate and memorable. 
\end{itemize}

\section{Related Works}
\subsection{Graph Learning}
Graph learning (CL) focuses on the information extraction from the non-Euclid data structure graph and accomplish target tasks such as node classification \cite{zhao2021graphsmote}, link prediction \cite{kazemi2018simple} and graph classification \cite{zhao2022consistency}.
Generally, a GL method requires analysis and combination of both the graph topology and node feature.
Most GL methods can be abstracted 
into a message-passing framework \cite{zhao2022topoimb,scarselli2008graph}, which is composed of two major operations: pattern extraction and aggregation within each layer \cite{zhao2022exploring}.
Numerous works have been conducted in exploring conventional based aggregation \cite{scarselli2008graph} or attention based aggregation \cite{velivckovic2017graph}.
However, these works are originally designed for static graphs and exhibit catastrophic forgetting problem when applied in OOD continual graph learning scenario.

\subsection{Continual Learning}
Our method also connects to continual learning \cite{de2021continual} (CL). 
CL studies the problem of learning from an infinite stream of data \cite{zhao2023towards}, with the goal of gradually extending acquired knowledge and adapt it to the future data or task \cite{zhao2020balancing,aljundi2019task,ren2018tracking}.
Current researches in continual learning could be classified into three major lines. 
\textit{Regularization-based methods} \cite{ahn2019uncertainty} mainly limit updates of important parameters for the prior tasks and penalize model parameters change by adding corresponding regularization terms \cite{zhao2018zero}.
\textit{Replay-based methods} store part of previous data in their raw format \cite{2018rawreplay} or pseudo-samples with a generative model \cite{shin2017replay}. 
\textit{Parameter isolation-based methods} \cite{zhou2021effective} dedicates different model parameters to different tasks to prevent the possible forgetting issue.

Recently, some emerging CL works have been applied to dynamic graph learning scenarios \cite{ego,zhao2022synthetic}. 
However, most of these works ignore graph intrinsic properties, 
and simply cast the continual graph learning into a classical CL problem \cite{ijcai2021traffic}. 
Differently, we dive into intrinsic properties of graphs and design a novel strategy by exploiting the message propagation w.r.t invariant latent relations behind edge existence, which can effectively adapt to evolving graphs.

\subsection{Unsupervised Domain Adaptation} 
Our work is relevant to unsupervised domain adaptation \cite{ZHOU2021domaiadpt,ren2022semi}. 
Unsupervised domain adaptation aims to transfer the knowledge learned from labeled source domains to the unlabeled target domain \cite{ren2021cross}. 
Existing works mainly focus on minimizing the discrepancy between the source and target distributions to learn domain-invariant features~\cite{tzeng2017adversarial,xu2018enhancing}.
Recently, some new works focus on domain adaptation on non-Euclid data, e.g., graph data \cite{domainCV}.
Yet, it is non-trivial to formulate graph distribution and design adaptive models considering its evolving topology, graph size and node features \cite{du2021dynamic,wu2022non}. 
One promising direction is to decouple the neighbor topology information from graphs and disentangle the original graph into a massive of ego-graph \cite{NEURIPS2021egodomain}. 
As a consequence, each ego-graph can be viewed as an i.i.d. sample and can be easily used for model training \cite{peng2022fedni}. 
However, node degrees can not be accurately estimated or used as prior knowledge, making the aforementioned works be sensitive to node degrees. Besides, these domain adaption works on graph data mainly consider static domains where only a source and a target domain are available. 
In contrast, our work targets on adaptive model training on sequential-evolving graph data. 


\section{Preliminary}

To formally define our shifted graph continual learning setting, 
we introduce a task incremental learning scenario $\mathcal{T}$ following~\cite{lifelong}, in which $N$ disjoint semi-node classification tasks $ \{ \tau_1, \tau_2,... , \tau_{N} \}$ evolve in sequence. 
Specificly,
each task $\tau_{t}$ is defined as:
$$\tau_{t} = \{ \mathcal{G}_t  , Y_t \}$$
where $\mathcal{G}_t=\{V_t,E_t,X_t \}$ denotes the undirected graph,
and $V_t = \{ v_t^{(1)} , v_t^{(2)},\cdots,v_t^{(n_t)} \}$ presents $n_t$ nodes. 
$E_t \subseteq \{ V_t \times V_t \}$ denotes the set of graph edges. 
The set $X_t = \{ x_t^{(1)} , x_t^{(2)},\cdots,x_t^{(n_t)} \}$ consists of the D-dimension node feature $x_t \in R^D$ for each node in the task $\tau_t$. 
Part of the nodes are labeled by $Y_t$ and denoted by $V_{t}^{tr}$. 
The other unlabeled nodes are denoted by $V_{t}^{se}$. 
Their node features are denoted as $X_t^{tr}$ and $X_t^{se}$ respectively.

Due to the data shift nature in the real world,
with the tasks performed successively, both the node feature and the graph structure shift. Besides, the label set $Y_t$ for each task $\tau_t$ is also different.
It is worth noting that the training process is conducted task by task and each task would only be used once for training.
Additionally, time step is visible both in the training and the testing stage.

In this work, we want to 
design a node classifier $C$, which could continually learn from the task sequence $\mathcal{T}$.
At any time $t$, the classifier $C$ is updated by the labeled data $\{V_t^{tr}, X_t^{tr},E_t\}$ and corresponding labels $Y_t$. The updated classifier is evaluated on the unlabeled data $\{V_t^{se}, X_t^{se}\}$.
An ideal target classifier not only adapts well to the new evolving task, but also avoids catastrophe forgetting, which means the model still performs well in the previous tasks.

\begin{figure*}[bt]
\vspace{-6mm}
\centering
\includegraphics[scale=0.6]{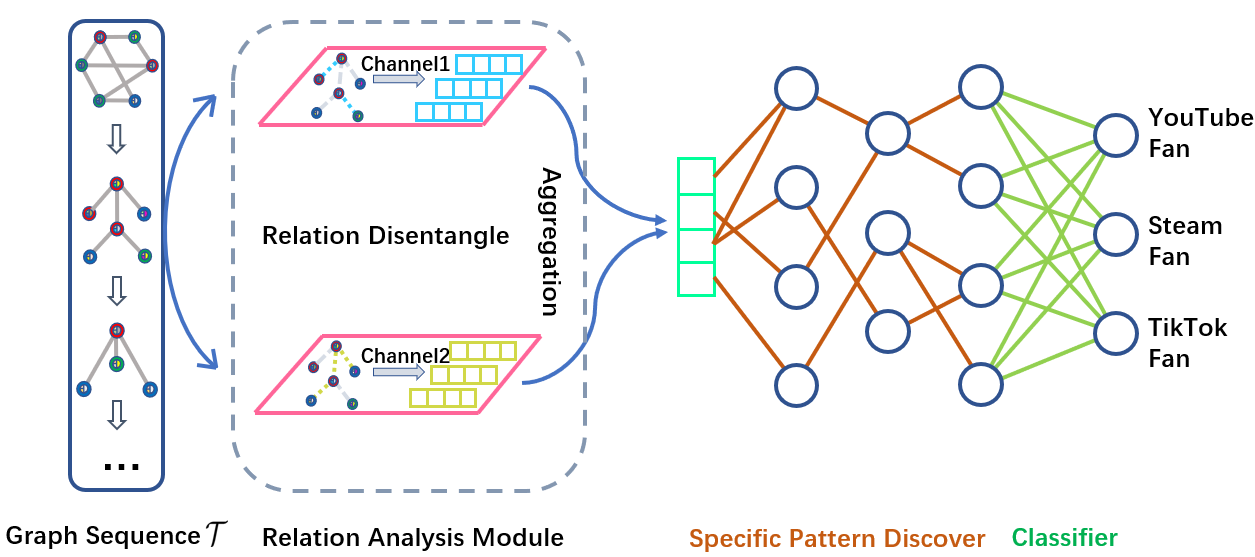}
\caption{An illustration of our RAM-CG method.}
\vspace{-4mm}
\label{model}
\end{figure*}

\section{Proposed Method}
In this section, we will elaborate the details of our RAM-CG. 
As shown in Figure \ref{model}, our proposed method consists of two modules, i.e., relation discovery module and task-awareness masking classifier.  
To capture the cross-domain knowledge, the relation discovery module implicitly extracts the latent relations behind the edges 
and integrates the relation information with the node features.
To improve the performance on each specific task and avoid forgetting the knowledge from previous tasks, 
we introduce task-awareness masking classifier, which learns a masking-based adaptive classifier.
We present our training algorithm in section A of the supplementary.

\subsection{Relation Discovery Module}
\noindent \textbf{Motivation}. 
We point out that the latent relations behind the edges could be attributed as an invariant factor for the evolving graphs. 
On the one hand, no matter how the distribution of nodes or edges is shifted, as the logic behind constructing edge to connect two nodes is stable, the cross-task knowledge does exist.
On the other hand, an edge in graphs presents a connection between two nodes, but the specific relation between the two nodes is seldom provided.
For example, graph constructed from social media generally building edge to abstract the follow relation between two users, but whether the follow relation coming from workmate relation or classmate relation is unclear. 
To improve the performance under OOD setting, 
we design the relation discovery module digging the cross-task knowledge by disentangling these latent relations.

The relation discovery module is a pyramid architecture like CNN for computer vision, where one CNN layer stacks on another CNN layer.
For simplicity, we only elaborate one relation discovery module for representing. 
The output of the previous module is the input of the next module.
Each relation discovery module consists of one relation disentangle layer and one relation aggregation layer.

\noindent \textbf{Relation Disentangle layer.}
The relation disentangle layer aims to implicitly extract the relations information via the designed multi-channel architecture. In each channel, node features are propagated along a specific relation implicitly. 
In specific, each channel calculates the propagation weights for every node with respect to each relation.
Concretely, the propagation weights between node $u$ and $v$ with respect to the $i$-th channel is denoted as:
\begin{align}
    \hat{e}^{i}_{u,v} = sigmod(W_2^{i} \cdot \sigma(W_1^{i} \cdot [h_u , h_v] )) \label{ eq:e } 
\end{align}
\begin{align}
    A_{u,v}^{i} = \frac{ exp( \hat{e}^{i}_{u,v} ) }{ \Sigma_{w \in \mathcal{U}} exp( \hat{e}^{i}_{u,w} ) } \label{ eq:A}
\end{align}
In equation (\ref{ eq:e }), $h_u$ and $h_v$ represent the input feature of node $u$ and node $v$. 
$[ h_u , h_v ]$ means that $h_u$ and $h_v$ are concatenated. 
$W_1^{i}$ and $W_2^{i}$ are two projection matrices. $\sigma$ is a nonlinear activation function, here we choose LeakyRelu. 
Following the previous work \cite{DBLP:journals/corr/KipfW16}, 
we conduct the normalization in the equation (\ref{ eq:A}), with the set $\mathcal{U}$ containing  all node $u$'s neighbors. $A_{u,v}^{i}$ denote the propagation weights between node $u$ and node $v$ under the relation $i$.

\noindent \textbf{Relation aggregation layer.} 
To generate new node embedding, 
the relation aggregation layer integrates the input node features and the invariant factor, i.e., the relations between two nodes.
In each channel, the node features are updated by the relation-aware propagation weights and neighbor nodes' features:
\begin{align}
    h_u^{i} = W_{feat}^{i}( \Sigma_{v \in \mathcal{U}} A_{u,v}^{i} \cdot  h_v  ) \label{new h}
\end{align}
where $h_v$ denotes the neighbor node of the node $u$, 
and $A_{u,v}^{i}$ is the relation-aware propagation weight between node $v$ and node $u$ with respect to channel (relation) $i$.
$W_{feat}^{i}$ is the corresponding projection matrix. 

To integrate the invariant information capturing from all latent relations, we aggregate the relation-aware feature learned by each channel via:
\begin{align}
    h_u^{'}= \sigma( W_{agg}^l [h_u^{1}, ... , h_u^{m+1}]  ) \label{h l+1}
\end{align}
where $h_u^{'}$ denotes the invariant features for node $u$ and 
$W_{agg}$ is the projection matrix.
$\sigma$ is a nonlinear activation function, here we choose LeakyRelu.

The final node feature from the relation discovery module in task $t$ is denoted as $h_t$.
Through stacking more modules, one node's feature is updated by further neighbors and relations.
However, too many modules may conduct an over-smoothing problem\cite{DBLP:journals/over/smooth} and reduce the model's discrimination ability.
To combine the above two sides, we stack two relation discovery module in this work.

\subsection{Task-awareness Masking Classifier }

The task-level relatively stable node relation only explains part of the invariant information across the shifted graphs, leaving the task-specific information unexplored.
To mine the variant information from $h_t$ and give the label prediction, we design the task-awareness masking classifier which is composed of a Specific Pattern Discover and an one layer MLP classifier.
The so called Specific Pattern Discover uses a Deep Neural Network $f$ as backbone following previous work in \cite{pmlr-v162-kang22b}. We note all the backbone parameters as $\theta$.

\noindent \textbf{Parameter Selection.}
Thanks to the Lottery Ticket Hypothesis \cite{DBLP:journals/corr/abs-1803-03635}, 
only a subset of the backbone $f$ could achieve a similar performance compared to the complete network. 
This hypothesis inspires us that we can select different groups of parameters from the backbone network $f$ to present the task specific knowledge.

To help accomplish parameter selection, we set a bunch of trainable score parameters $\mathcal{S}$, with each score parameter judging the importance of each parameter in the backbone one to one.
A parameter in backbone $f$ with higher absolute score value would be considered as a key parameter in pattern discover.
For each time $t$, these score parameters $\mathcal{S}$ would be updated every training loop.
After the training stage, we note the backbone's parameters with the highest top $c\%$ absolute score as $\theta_t$.
When testing on task $t$, we will activate them with a parameter mask.
The union set of selected parameters of all tasks is denoted as $U_t = \bigcup_{i=1}^{t} \theta_i $.
The selection ratio $c\%$ will be discussed in Sec.~\ref{sec_exp}.

\noindent \textbf{Inference and Update.}
When the new task evolves at time $t+1$, 
during each forward training stage, we use the parameter selection method to activate a subset of the backbone $f$'s parameters, which is denoted as ${\phi}_{t+1}$. Concretely, if the score parameter of one parameter is larger than a pre-defined threshold, the parameter will be selected. 
Those selected parameters construct the sub-network $f_{\phi_{t+1}}$ for the new task.  
And then we leverage the one layer MLP classifier $g$ to classify nodes. 

The inference of model is formulated as:
\begin{align}
    \hat{Y}_{t+1} = g(f_{\phi_{t+1}}(h_{t+1}))
\end{align}
where
$h_{t+1}$ denotes the invariant node features for the $t+1$ task from the  relation discovery module. 


The objective function is the node classification error:
\begin{align}
    \mathcal{L} = nll\_loss(log(\hat{Y}_t),Y_t)
\end{align}

In the backward stage, only the parameters $\xi_{t+1}=\phi_{t+1} \bigcap U_t^{c}$ will be updated by the current task while the other parameters are fixed to preserve the knowledge learned from previous tasks. 
Inside the optimizer, we update $\xi_{t+1}$ as:
\begin{align}
    \xi_{t+1} \gets \xi_{t+1} - \eta ( \frac{\partial \mathcal{L}}{\partial \xi_{t+1}}) \label{up th ta1}
\end{align}


We also update the score parameters $\mathcal{S}$ by the loss and find better score parameters for all tasks:
\begin{align}
    \mathcal{S} \gets \mathcal{S} - \eta \cdot \frac{\partial \mathcal{L}}{\partial \mathcal{S}}
\end{align}
Via reusing the parameters, the cross domain knowledge will be shared over all tasks.
{For any sub-parameter-set of $U_t$, a high score value of it would encourage a higher overlap between $U_t$ and $\theta_{t+1}$.}

\subsection{Training Strategy}

{As the size of labeled nodes is much larger for the first task and the latent relation discovery is invariant across tasks, we train the relation analysis module only on the first task which shall be sufficient.
We also freeze the classifier after the first task, preserving a stable latent labeling function space.
Such that, the specific pattern discover is forced to capture the task-specific knowledge  with discovered relations in a self-modulation manner. }

\section{Experiments}
\label{sec_exp}
We conduct extensive experiments on a variety of datasets to evaluate the performance of our method with the following question:

$\bullet$ \textbf{Q1:} How does the proposed RAM-CG perform?

$\bullet$ \textbf{Q2:} How do the sensitive hyper parameters affect RAM-CG?

$\bullet$ \textbf{Q3:} How do the latent relations contribute to the success of RAM-CG?

$\bullet$ \textbf{Q4:} What is the procession detail of the RAM-CG?

\begin{table}[t]
    \centering
    \begin{tabular}{lcccc}
        \toprule

        Dataset  &CitationNet  & OGBN-arxiv & TWITCH \\
        
        \midrule
        \# node     & 23779  &  344252    &  34120  \\
        \# edge   &   38570 &    1166243  &  429113  \\
        \# features & 6775 & 128  & 3169 \\
        \# classes  & 6 & 40  & 2 \\
        \# tasks    & 6 & 7  & 6 \\
      
        \bottomrule
    \end{tabular}
    \caption{Statics of Datasets}
    \label{tab:booktabs}
\end{table}

\begin{figure*}[bt]
\centering	
	\begin{minipage}{0.3\linewidth}
		\vspace{3pt}
		\centerline{\includegraphics[width=\textwidth]{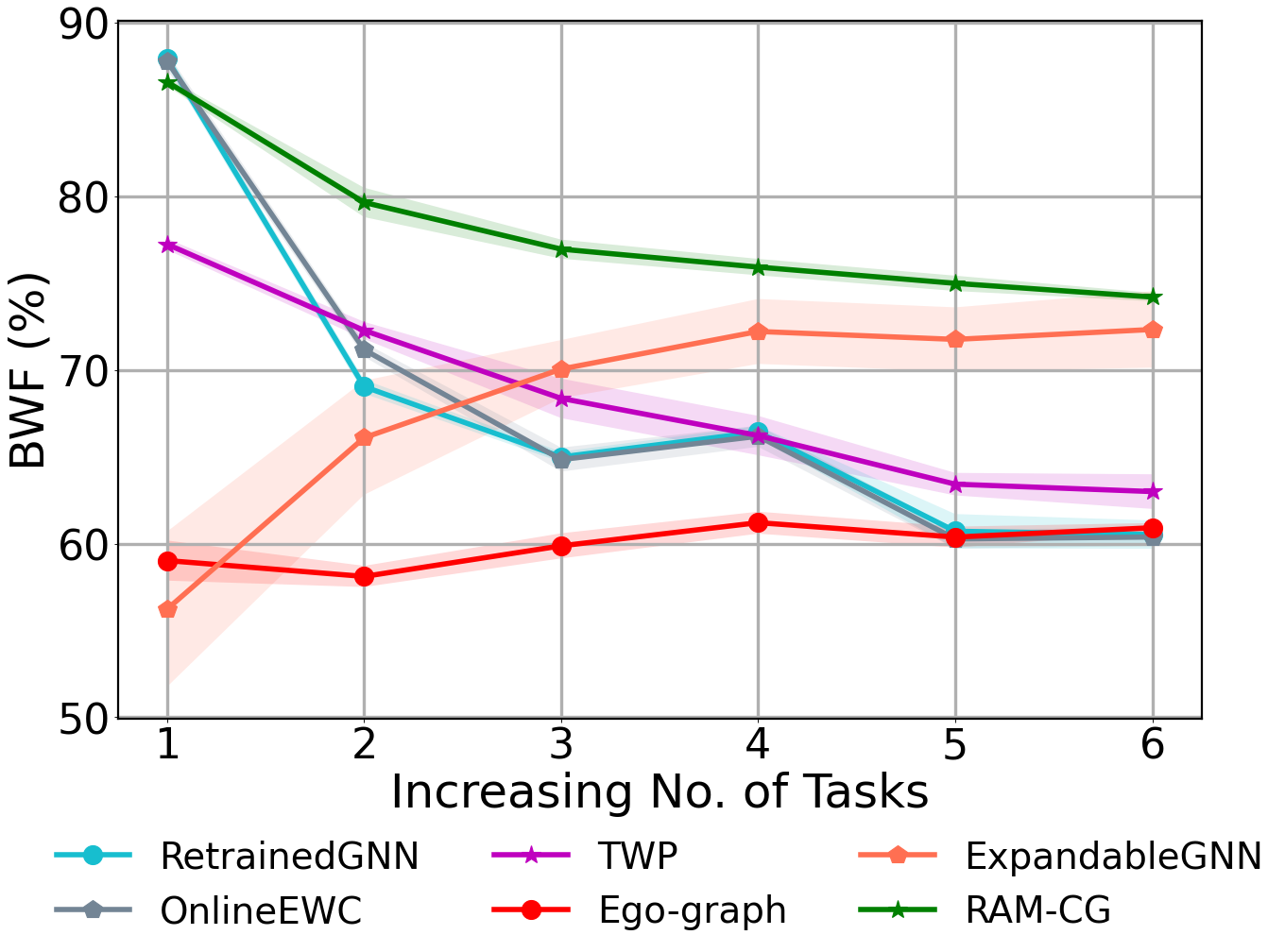}}
	\end{minipage}
	\begin{minipage}{0.30\linewidth}
		\vspace{3pt}
		\centerline{\includegraphics[width=\textwidth]{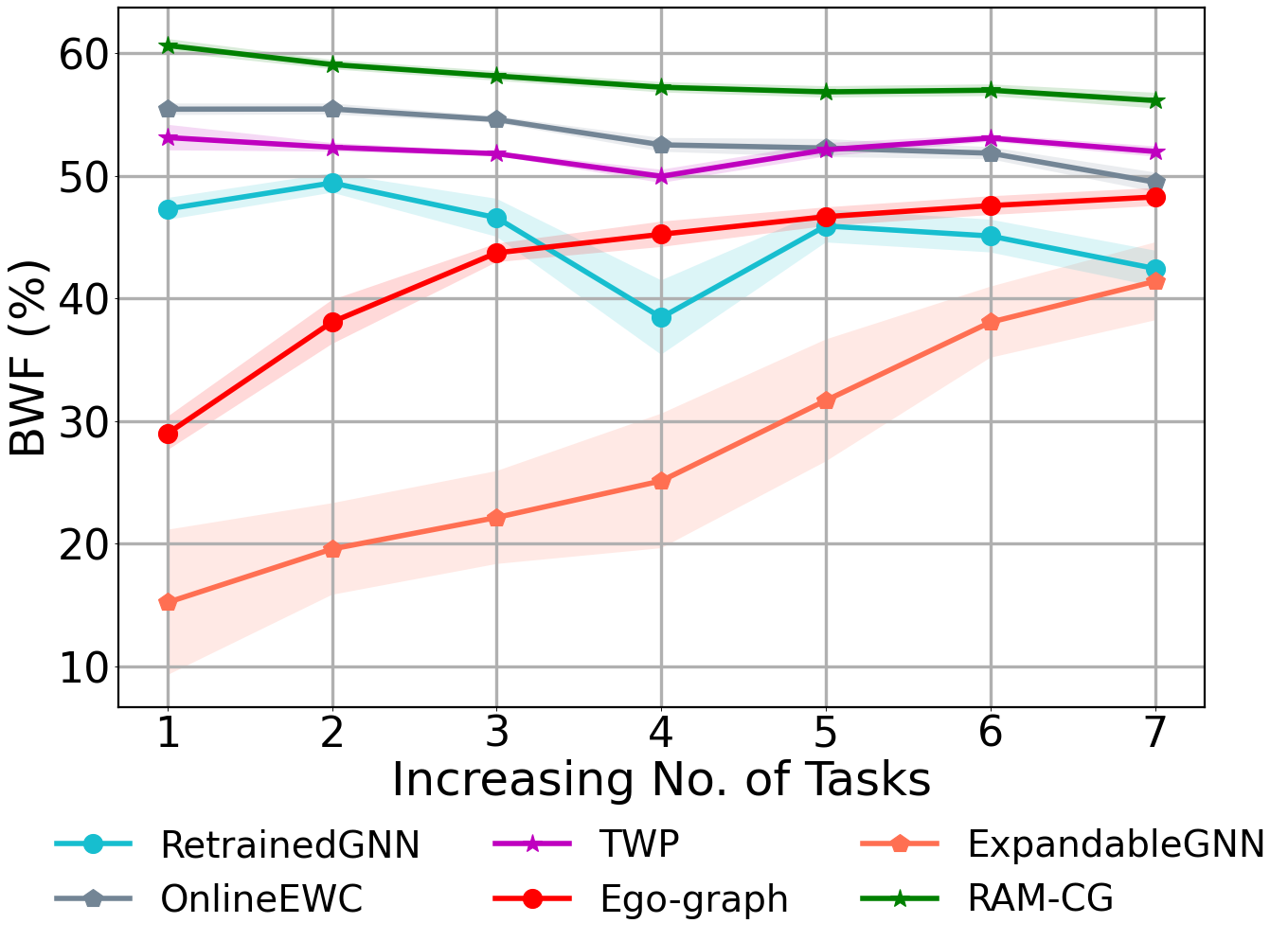}}
	 
	\end{minipage}
        \begin{minipage}{0.30\linewidth}
		\vspace{3pt}
		\centerline{\includegraphics[width=\textwidth]{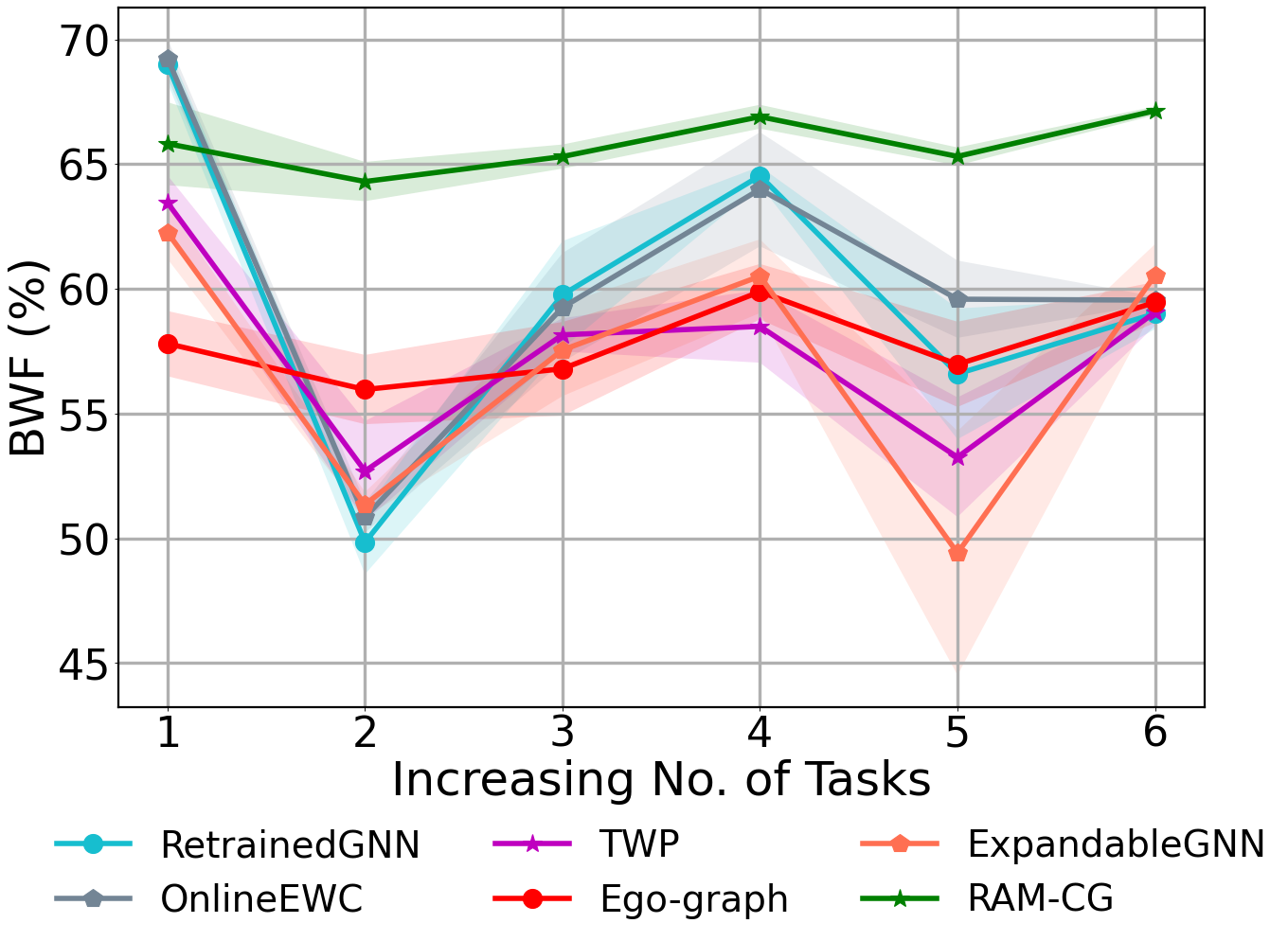}}
	 
	\end{minipage}

        \begin{minipage}{0.30\linewidth}
		\vspace{3pt}
		\centerline{\includegraphics[width=\textwidth]{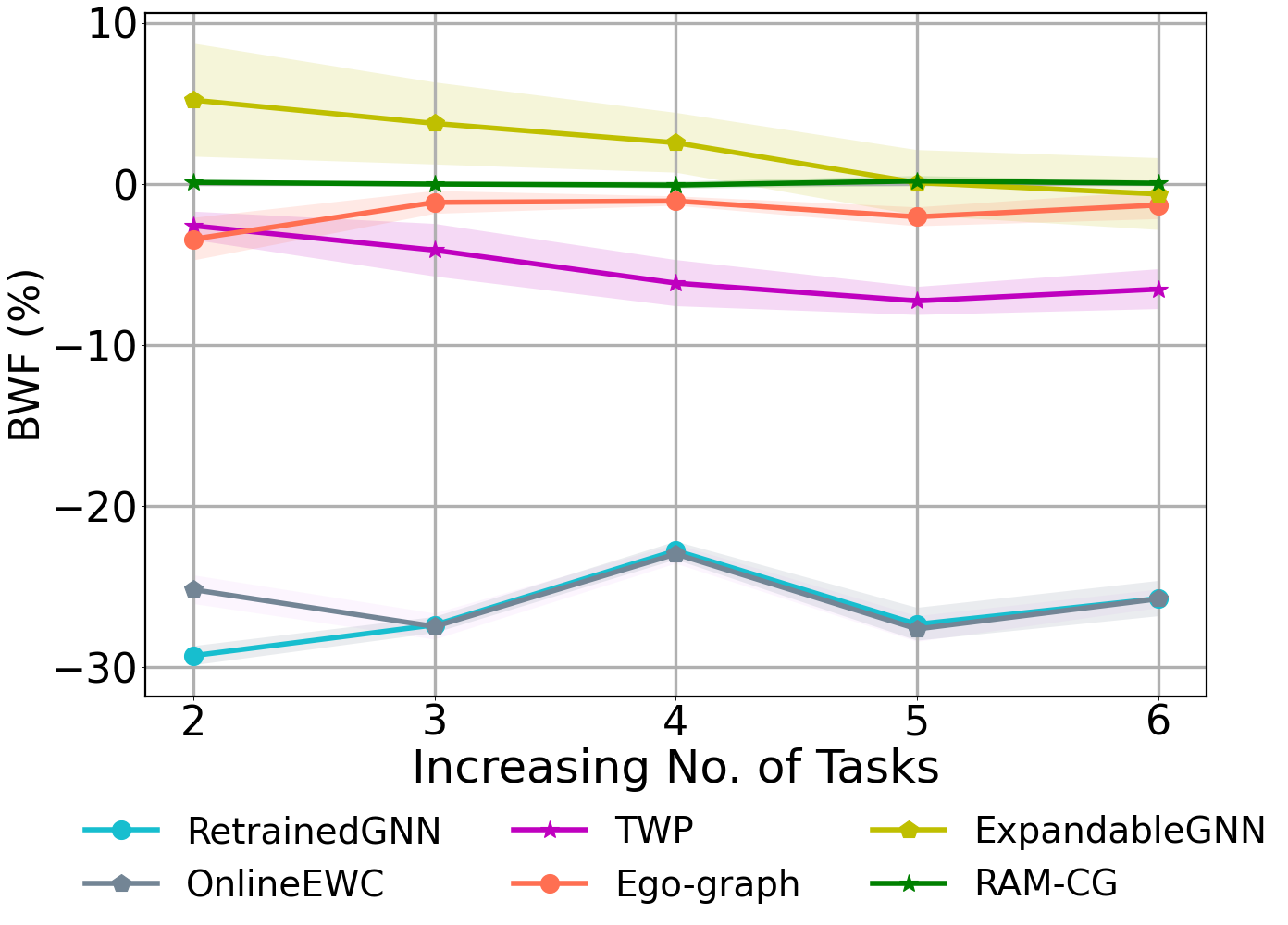}}
		\centerline{CitationNet}
	\end{minipage}
	\begin{minipage}{0.30\linewidth}
		\vspace{3pt}
		\centerline{\includegraphics[width=\textwidth]{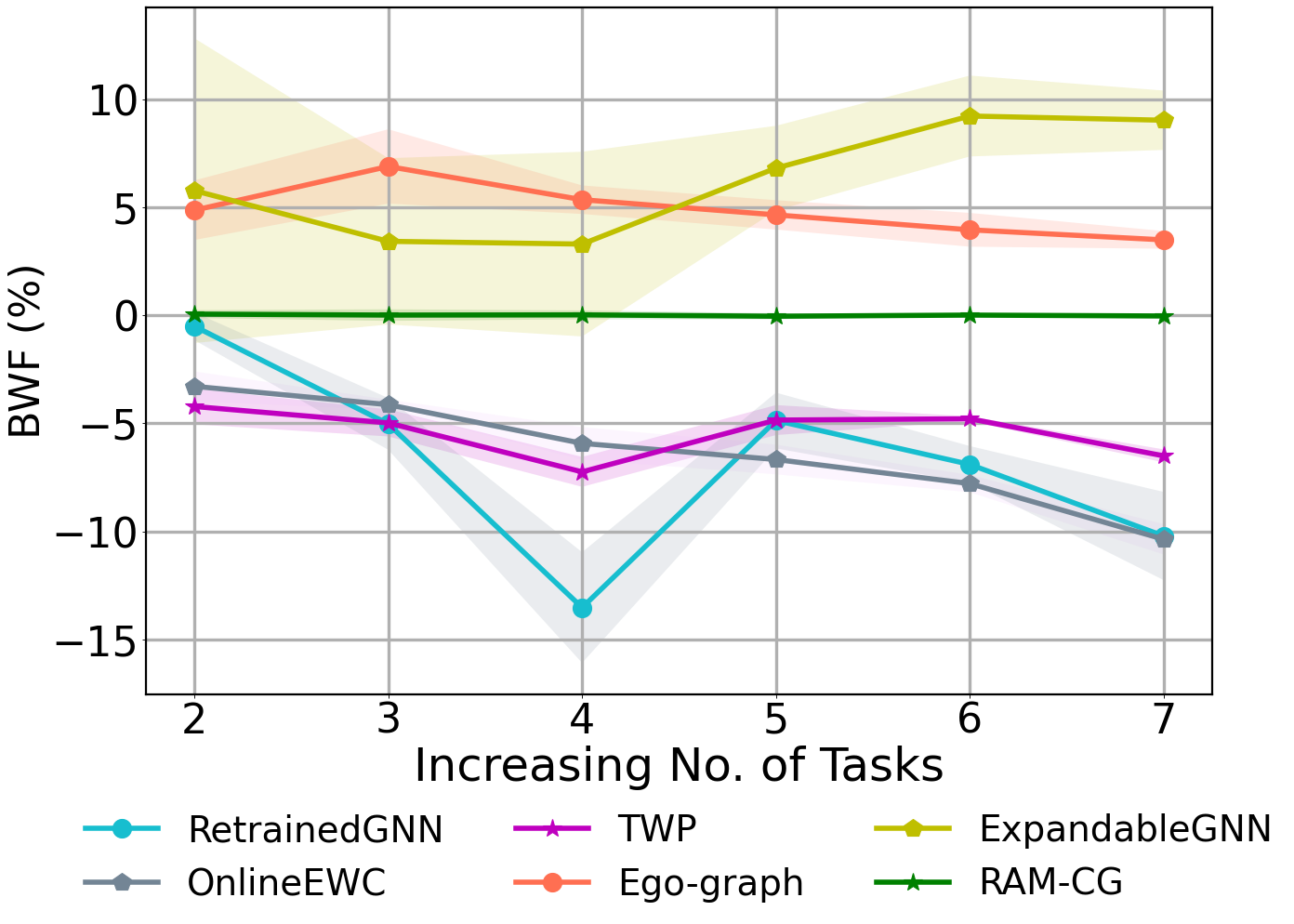}}
	 
		\centerline{OGBN-arxiv}
	\end{minipage}
        \begin{minipage}{0.30\linewidth}
		\vspace{3pt}
		\centerline{\includegraphics[width=\textwidth]{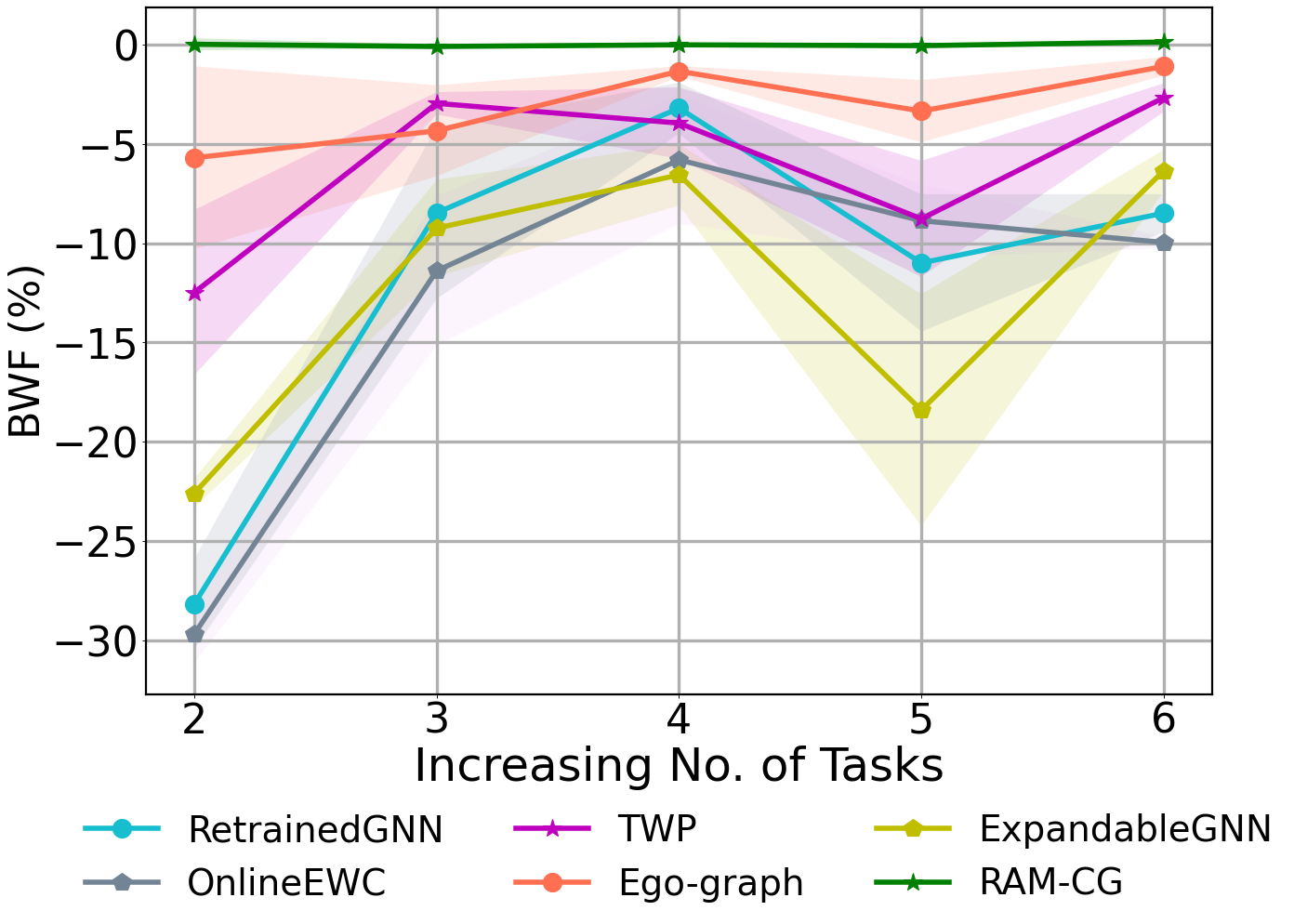}}
	 
		\centerline{TWITCH}
	\end{minipage}

	\caption{ The overall performance }
	\label{pic:overall}

\end{figure*}

\subsection{Experiments Setup}
\textbf{Datasets.} We build three OOD incremental graph node classification datasets based on three graph datasets.
In this section, we will briefly introduce those three graph datasets. Table \ref{tab:booktabs} shows the statistical summary of the datasets. The details of how to reconstruct the datasets will be elaborated in section B.1 of the supplementary.

$\bullet$ CitationNet. 
We choose three citation network `citation v1', `dblp v7' and `acm v9' from ArnetMiner \cite{Tang:08KDD}.

$\bullet$ OGBN-arXiv.
Researchers collect papers in forty fields of Computer Science from arXiv across decades of years \cite{10.1162/qss_a_00021}.
We select papers from different time periods and use them to build graphs with the papers they cite.

$\bullet$ TWITCH.
The TWITCH dataset consists of users' social networks from different regions in the Twitch platform \cite{rozemberczki2021twitch}.

{For each dataset, we take 60\% nodes for train stage, 20\% for validation stage and 20\% for test stage in the first task, while 30\% nodes for train stage, 20\% nodes for validation stage and 50\% nodes for test stage in the other tasks.} \\

\textbf{Evaluation Metrics.} We denote the test accuracy on data set $V_j^{se}$ after training on data set $V_i^{tr}$ as $R_{i,j}, i \leq j$.
Following general continual learning frameworks, at any time $T$,
we introduce Average Accuracy(ACC):  
\begin{align}
    ACC_T = \frac{1}{T} \sum_{t=1}^{t=T} R_{T,t}
    \label{ACC}
\end{align}
e.g. the average performance among all evolved tasks, 
to judge model's ability to make prediction.

To measure the memorization ability of different method, we use the Backward Transfer (BWF) as an evaluation metric: 
\begin{align}
    {BWF}_T = \frac{1}{T-1} \sum_{t=1}^{t=T-1} (R_{T,t}-R_{t,t})
    \label{BWF}
\end{align}
e.g. the average performance variation in previous task after training on new task,
to measure the possible model forecast ability decrease.
We note that BWF could be a positive number, which means the new knowledge model learned improves the prediction for old tasks.\\

\textbf{Baselines.}
We compare our RAM-CG method with six competitive baselines.

$\bullet$ RetrainedGNN.
We directly train a GCN model on the training set task by task without any external measures against catastrophic forgetting.

$\bullet$ JointGNN.
We train a GCN model on the joint training set of each dataset and evaluate it on each test set.
We take it as an estimated \textit{upper bound} for ACC.

$\bullet$ Online Elastic Weight Consolidatiokn (OnlineEWC).
OnelineEWC \cite{pmlr-v80-schwarz18a} is a general regularization based continual learning method, which penalizes the change of parameters compare to the current ones when a new task comes.

\begin{table*}[t]
\caption{Metric in the last step}

\resizebox{\linewidth}{!}{
\begin{tabular}{|c|cc|cc|cc|cc}

\hline

\multirow{2}{*}{Method} & \multicolumn{2}{c|}{CitationNet} & \multicolumn{2}{c|}{OGBN-arxiv}   & \multicolumn{2}{c|}{TWITCH}    \\
 & ACC (\%) & BWF (\%) & ACC (\%) & BWF (\%)  & ACC (\%) & BWF (\%)    \\

\hline

\small{RetrainedGNN}
& \footnotesize{60.52}\scriptsize{$\pm$0.81}
&-25.75\scriptsize{$\pm$1.09}

&\footnotesize{42.44}\scriptsize{$\pm$1.46}
&\footnotesize{-10.21}\scriptsize{$\pm$2.04}

&\footnotesize{59.00}\scriptsize{$\pm$0.62}
&\footnotesize{-8.49}\scriptsize{$\pm$0.95}\\

\small{OnlineEWC}
&\footnotesize{60.38}\scriptsize{$\pm$0.51}
&\footnotesize{-25.77}\scriptsize{$\pm$0.57}

&\footnotesize{49.48}\scriptsize{$\pm$0.79}
&\footnotesize{-10.36}\scriptsize{$\pm$0.70}

&\footnotesize{59.54}\scriptsize{$\pm$0.14}
&\footnotesize{-9.96}\scriptsize{$\pm$0.30}\\

\small{TWP}

&\footnotesize{63.00}\scriptsize{$\pm$0.99}
&\footnotesize{-6.53}\scriptsize{$\pm$1.23}

&\footnotesize{51.98}\scriptsize{$\pm$0.42}
&\footnotesize{-6.50}\scriptsize{$\pm$0.30}

&\footnotesize{59.06}\scriptsize{$\pm$0.54}
&\footnotesize{-2.68}\scriptsize{$\pm$0.72}\\

\small{Ego-graph} 

&\footnotesize{60.92}\scriptsize{$\pm$0.27}
&\footnotesize{-1.31}\scriptsize{$\pm$0.87}

&\footnotesize{48.27}\scriptsize{$\pm$0.73}
&\footnotesize{3.48}\scriptsize{$\pm$0.40}

&\footnotesize{59.45}\scriptsize{$\pm$0.82}
&\footnotesize{-1.11}\scriptsize{$\pm$0.46}\\

\small{ExpandableGNN}

&\footnotesize{72.00}\scriptsize{$\pm$1.66}
&\footnotesize{-0.62}\scriptsize{$\pm$2.02}

&\footnotesize{41.39}\scriptsize{$\pm$4.97}
&\footnotesize{9.02}\scriptsize{$\pm$1.87}

&\footnotesize{60.54}\scriptsize{$\pm$1.47}
&\footnotesize{-6.35}\scriptsize{$\pm$5.83}\\

\small{RAM-CG}
&\footnotesize{\bf{74.21}}\scriptsize{$\pm$0.25}
&\footnotesize{0.04}\scriptsize{$\pm$0.16}

&\footnotesize{\bf{56.13}}\scriptsize{$\pm$0.64}
&\footnotesize{-0.03}\scriptsize{$\pm$0.04}

&\footnotesize{\bf{67.13}}\scriptsize{$\pm$0.17}
&\footnotesize{0.11}\scriptsize{$\pm$0.15}\\

\hline
\small{JointGNN}
&\footnotesize{80.86}\scriptsize{$\pm$0.16}
& -

&\footnotesize{57.90}\scriptsize{$\pm$0.01}
&-

&\footnotesize{66.31}\scriptsize{$\pm$0.66}
& - \\

\hline

\end{tabular}
}
\label{Tab:forget_competence}
\end{table*}

$\bullet$ Topology-aware Weight Preserving (TWP).
TWP \cite{TWP} is a plug-and-play continental learning algorithm specially designed for graph-continue problems.
The topological information of graphs is under concern, and the important parameters will be limited variance to store old knowledge when learning a new task.

$\bullet$ Ego-graph Replay.
Based on the discovery that ego-graphs(ego-networks) can provide locality-sensitive information, \cite{ego} proposes an ego-graph replay based continual learning algorithm.
Some ego graphs are sampled when new tasks evolve and make up a reply database.

$\bullet$ ExpandableGNN.
We introduce the ExpandableGNN from \cite{ExpandableGNN}, which contains a reinforcement learning based controller network to conduct a neural architecture search (NAS) for decisively selecting neural network components from a predefined search space.\\
\textbf{Model Configuration.}
When conducting experiments on the CitationNet dataset with RAM-CG, we stack two analysis modules with six relation disentangle channels and set the parameter selection ratio to 0.7 for task-awareness masking classifier. 
On the other two datasets, we use relation discovery module with four relation disentangle channels and keep other settings.
We take an Adam optimizer to train the model with a learning rate of 0.0065.

\subsection{Over all Comparison}
To answer Q1, we compared our RAM-CG with baselines.
Figure \ref{pic:overall} shows the average accuracy (ACC) in the first line and backward transfer (BWF) in the second line on the test data across the evolving tasks for four runs. 
We calculate the evaluation metrics, i.e., ACC and BWF, after learning each new evolving task, which represents the standard error computed from four runs with random seeds by the shaded regions. 
The ACC curves of our model are more gradual than other baselines. Besides, the curves of other methods distinctly shake.
These figures confirm that our model could learn from a graph sequence more robustly. 
Meanwhile, the BWF curve of our model slightly oscillates around the zero axis, which means our RAM-CG model successfully avoids catastrophe forgetting.

Table \ref{Tab:forget_competence} shows the ACC and BWF after learning the last task.
The retrainedGNN acts out catastrophic forgetting on all three datasets, implying the existence of sequential graph shift.
The ACC of our model outperforms the ACC of baselines by a large margin of $2.2\%$  on CitationNet, $4.2\%$ on OGBN-arxiv and $6.6\%$ on TWITCH.
Our RAM-CG even performs better than the JointGNN on TWITCH dataset by 0.8 point, demonstrating that our model has great advantages in consistent learning from sequential data. 
At the same time, our model effectively alleviates catastrophic forgetting and even promote BWF to a positive number in CitationNet and OGBN-arxiv dataset. 
Meanwhile, the ego-graph replay method has a better knowledge transfer ability, for the direct expansion of knowledge sources.

\subsection{Hyper Parameter Analysis}
\begin{figure*}[bt]
\centering	
	\begin{minipage}{0.24\linewidth}
		\vspace{3pt}
		\centerline{\includegraphics[width=\textwidth]{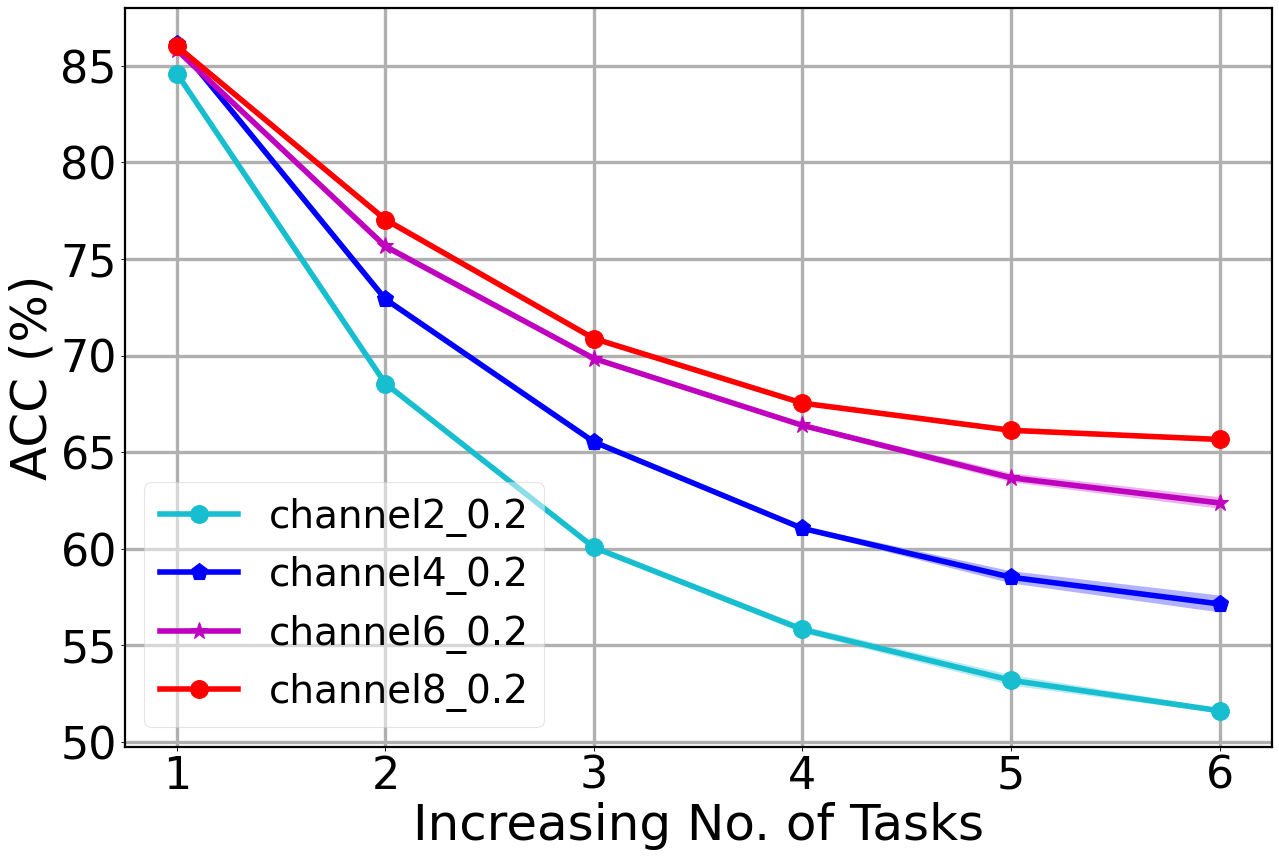}}
		\centerline{ratio = 0.2}
	\end{minipage}
        \begin{minipage}{0.24\linewidth}
		\vspace{3pt}
		\centerline{\includegraphics[width=\textwidth]{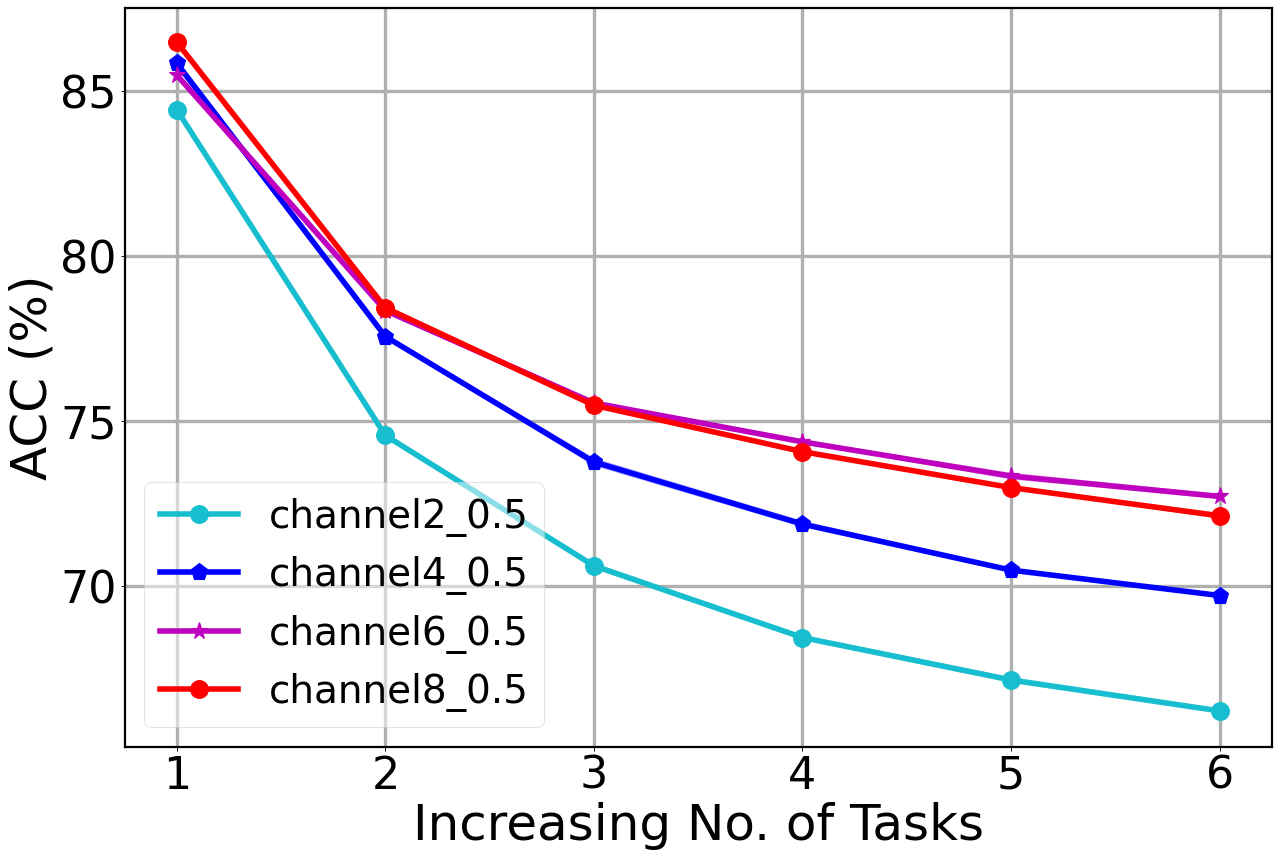}}
	 
		\centerline{ratio = 0.5}
	\end{minipage}
	\begin{minipage}{0.24\linewidth}
		\vspace{3pt}
		\centerline{\includegraphics[width=\textwidth]{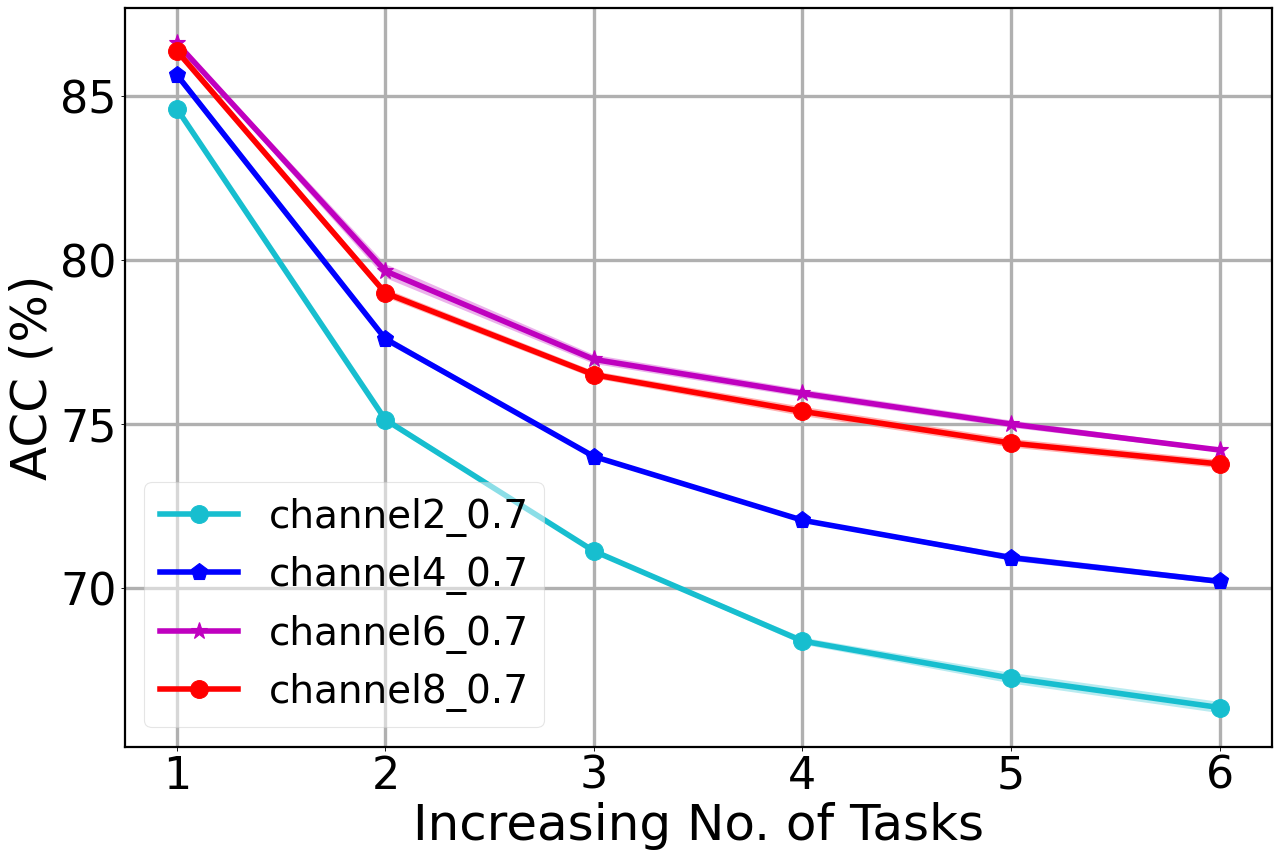}}
	 
		\centerline{ratio = 0.7}
	\end{minipage}
        \begin{minipage}{0.24\linewidth}
		\vspace{3pt}
		\centerline{\includegraphics[width=\textwidth]{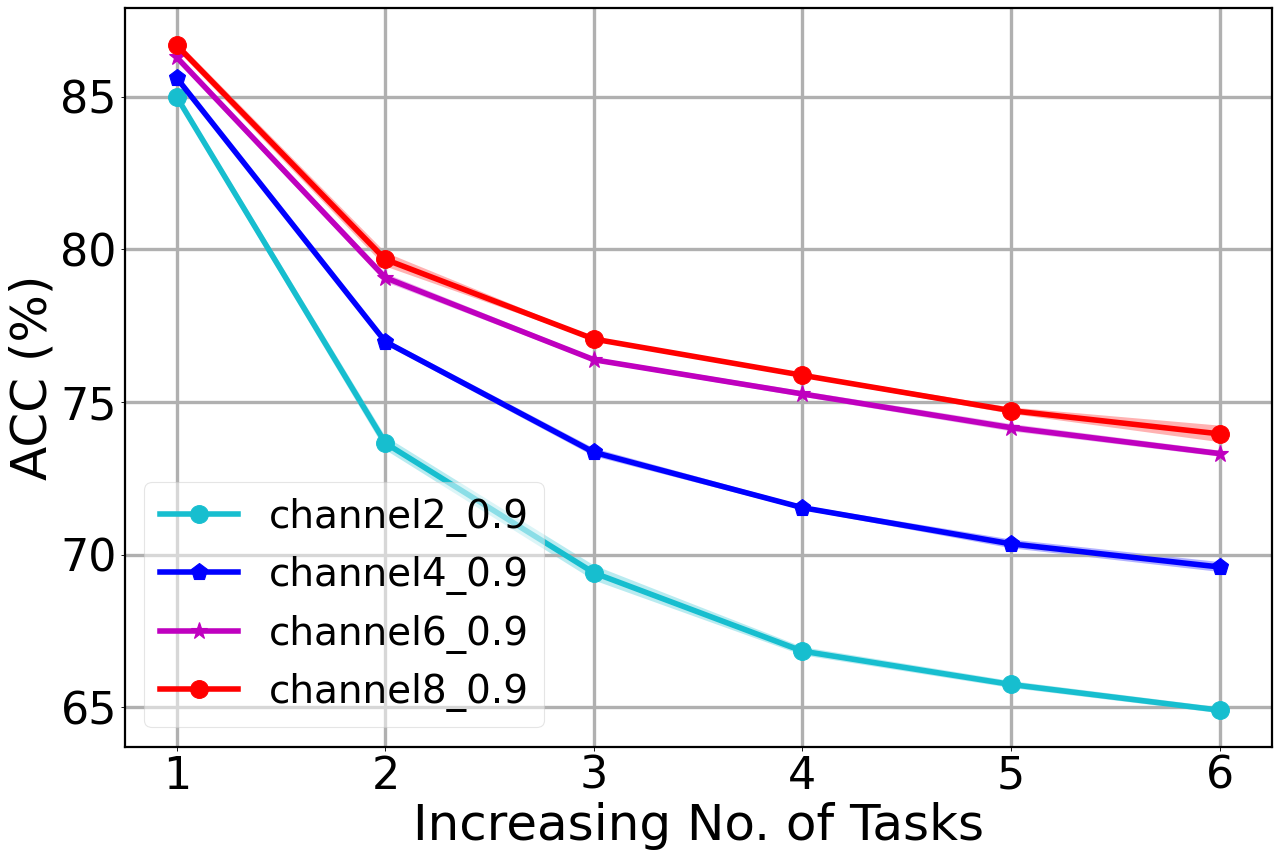}}
	 
		\centerline{ratio = 0.9}
	\end{minipage}

        \begin{minipage}{0.24\linewidth}
		\vspace{3pt}
		\centerline{\includegraphics[width=\textwidth]{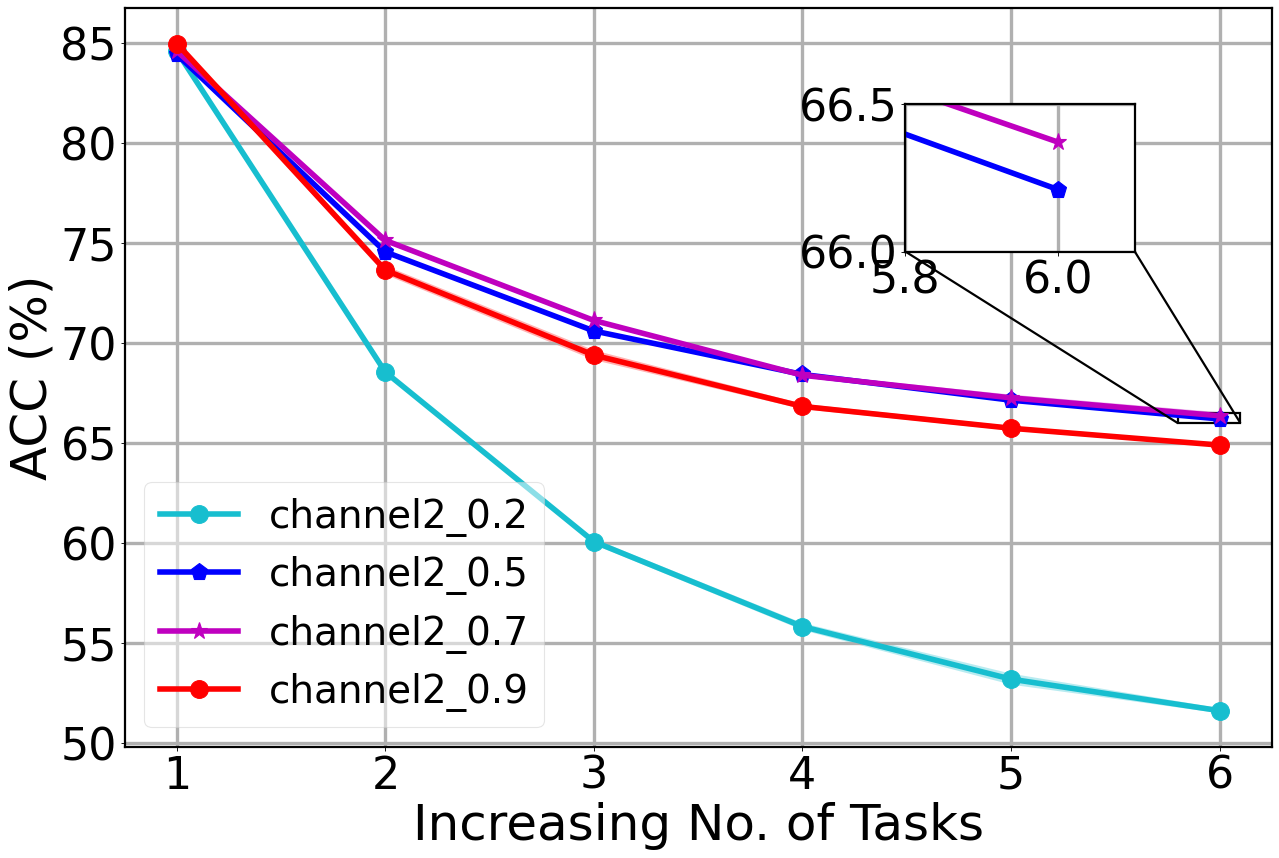}}
		\centerline{channel = 2}
	\end{minipage}
        \begin{minipage}{0.24\linewidth}
		\vspace{3pt}
		\centerline{\includegraphics[width=\textwidth]{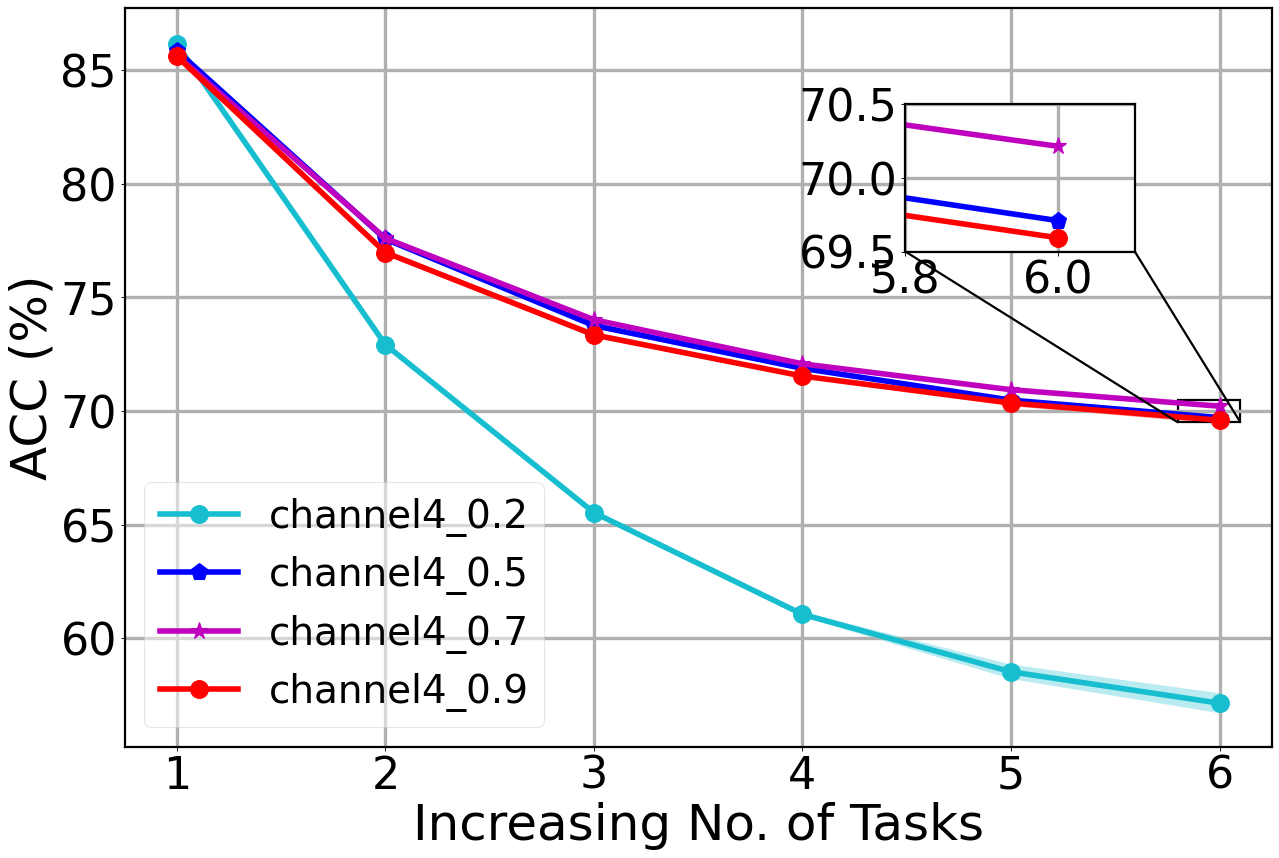}}
	 
		\centerline{channel = 4}
	\end{minipage}
	\begin{minipage}{0.24\linewidth}
		\vspace{3pt}
		\centerline{\includegraphics[width=\textwidth]{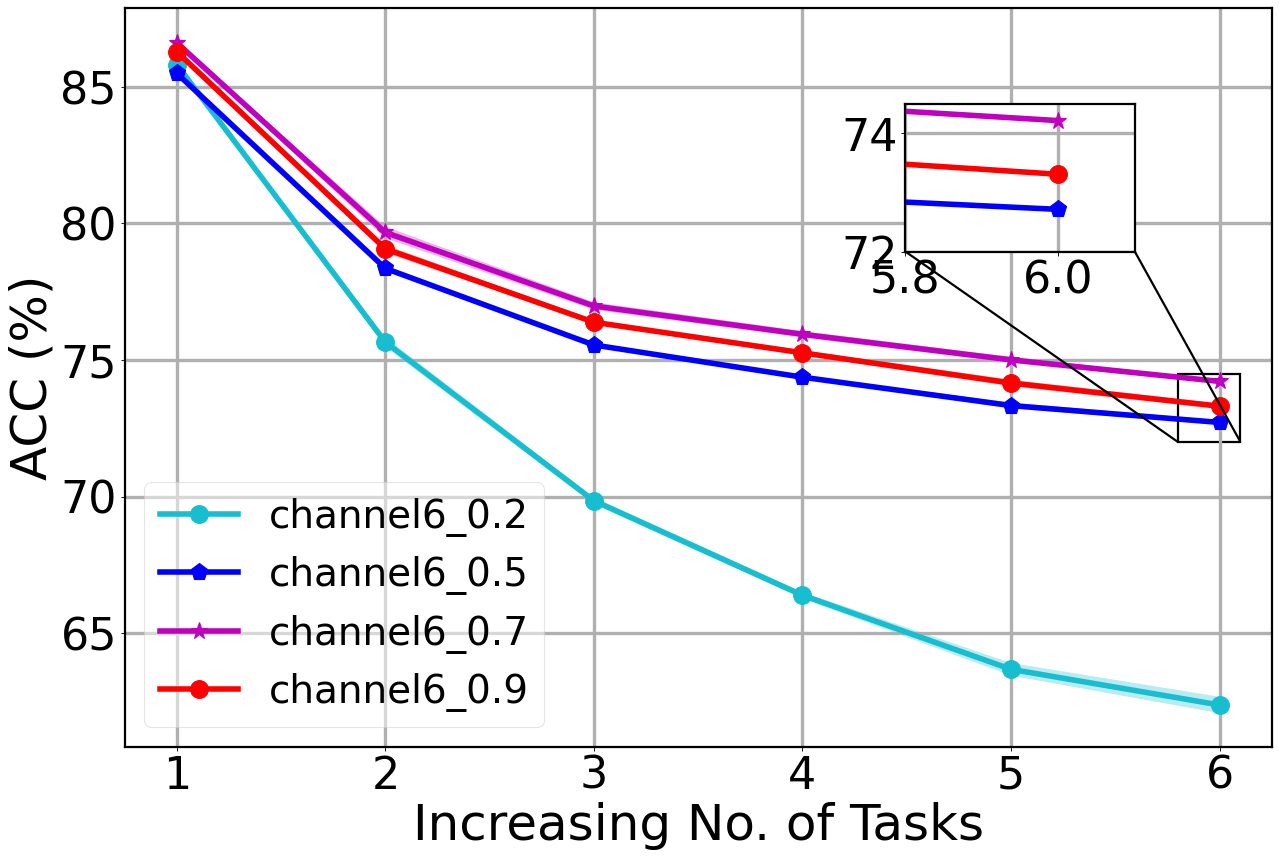}}
	 
		\centerline{channel = 6}
	\end{minipage}
        \begin{minipage}{0.24\linewidth}
		\vspace{3pt}
		\centerline{\includegraphics[width=\textwidth]{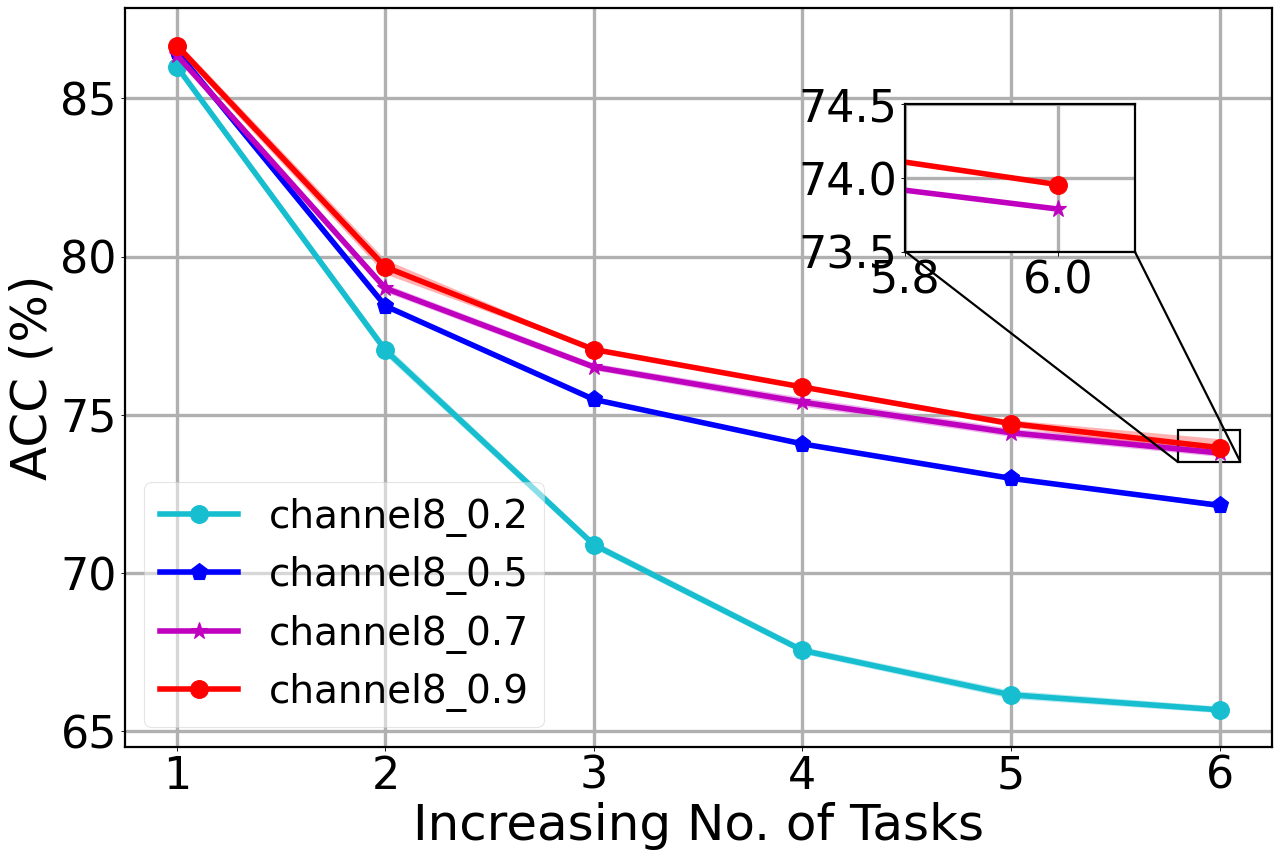}}
	 
		\centerline{channel = 8}
	\end{minipage}
 
	\caption{ Hyper Parameter Sensitive  Study }
	\label{pic:sensitive}

\end{figure*}

To answer Q2, we study two key hyper parameters: the channel heads number of relation discovery module and the parameter selection ratio $c\%$ in task-awareness masking classifier modeling on the CitationNet dataset.
We permute four candidate channel head choices: 2, 4  6 and 8, with four parameter selection ratios: 0.2, 0.5, 0.7 and 0.9.

As shown in Figure \ref{pic:sensitive}, we compare the model performance with a fixed parameter selection ratio in the first line and fixed channel heads in the second line.
As using more channel heads enables the model to distinguish more latent relations, when we fix the parameter selection ratio, the model with 6 channel heads apparently performs better than the model with 2 or 4 channel heads.
However, except the parameter selection ratio equals 0.2, the difference between the model with 6 channel heads and the model with 8 channel heads is slight.
Because the amount of latent relations is an intrinsic characteristic of a dataset, the promotion of add channel heads would finally meet an upper bound.
{Meanwhile, the influence of increasing parameter selection ratio appears a peak. 
The small ratio model is apparently worse than the large one, while the other models get similar performance, but the largest ratio models don't outperform.}
In fact, if more parameters are selected for one task, the potential parameter space for rest tasks would be reduced.
The decision of parameter selection ratio appears a trade-off pattern.

\subsection{Ablation Study}

\begin{figure}[bt]
\centering	
	\begin{minipage}{0.49\linewidth}
		\centerline{\includegraphics[width=\textwidth]{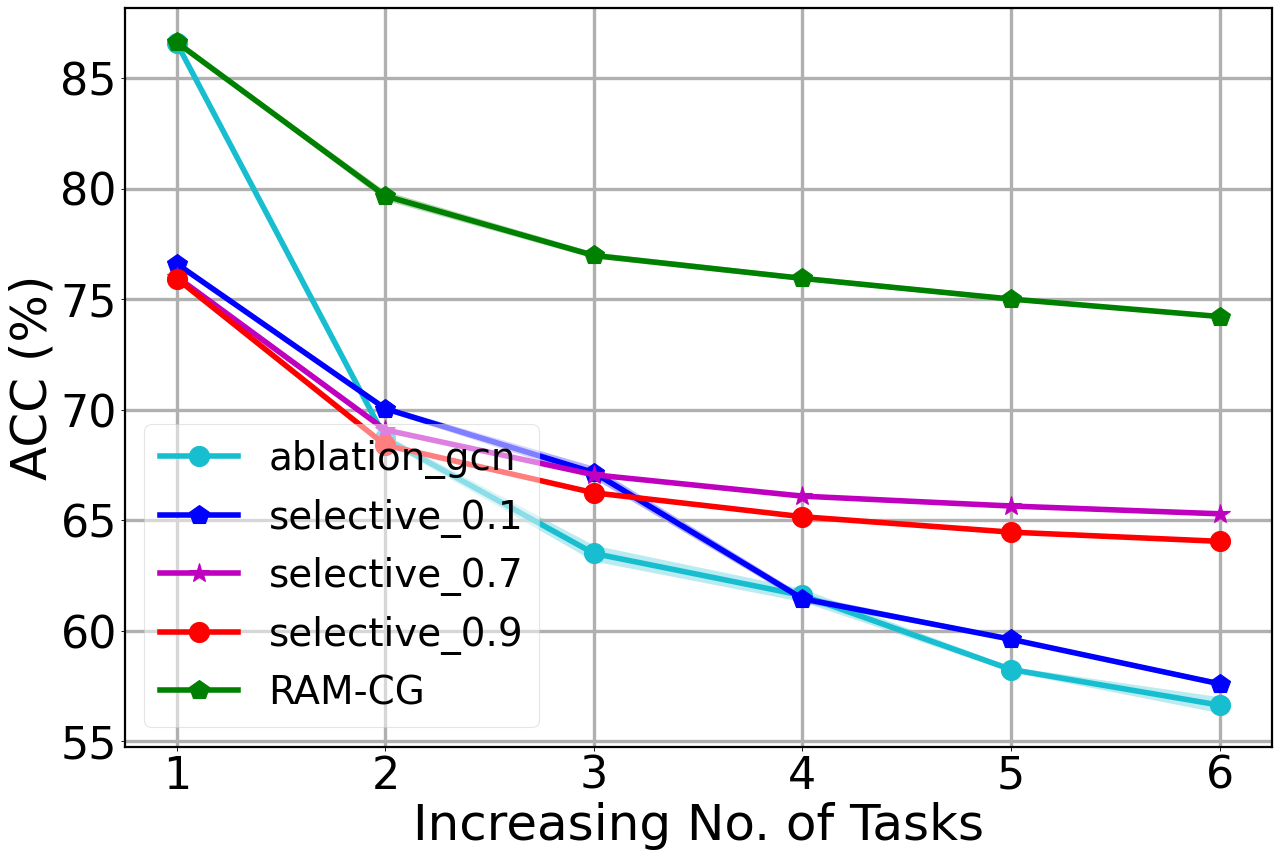}}
		\centerline{CitationNet}
	\end{minipage}
        \begin{minipage}{0.49\linewidth}
		\centerline{\includegraphics[width=\textwidth]{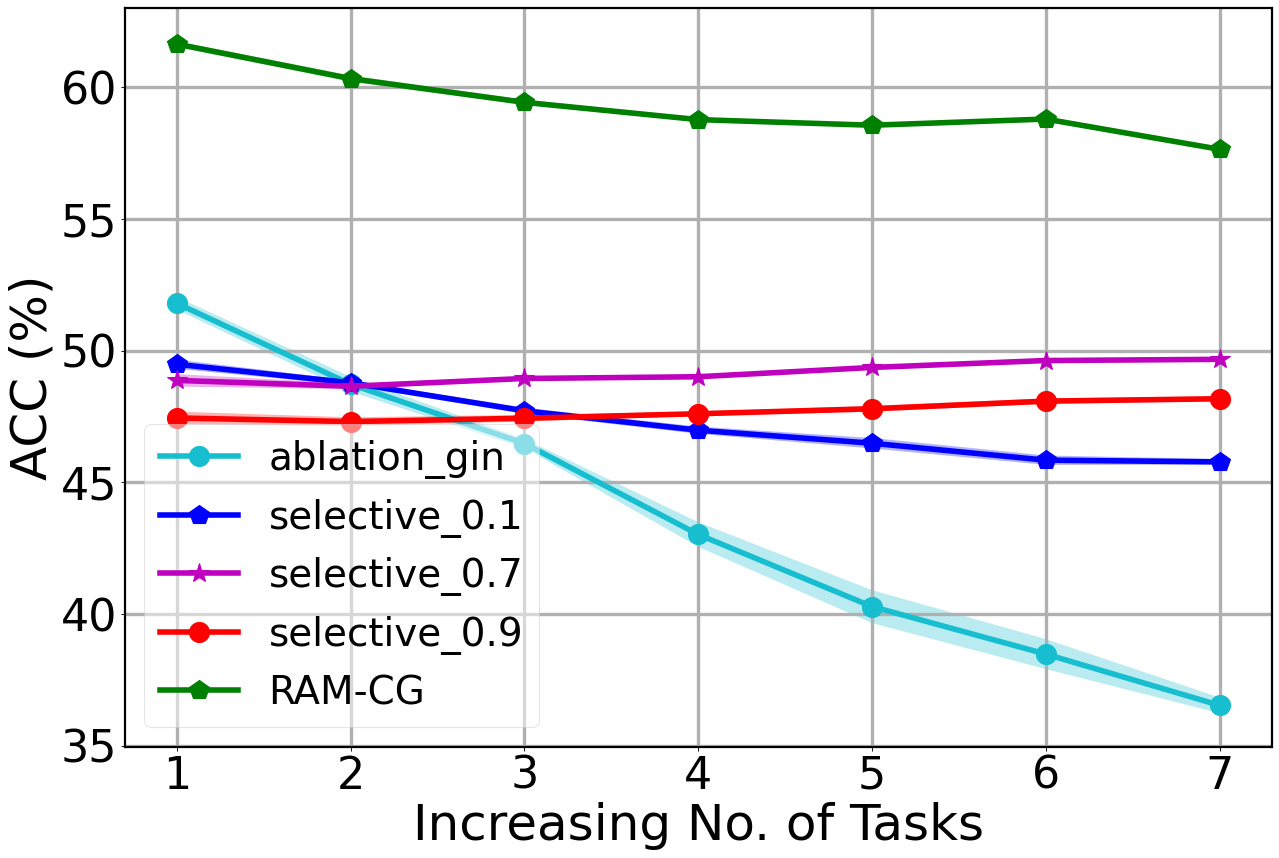}}
	 
		\centerline{OGBN-arxiv}
	\end{minipage}

	\caption{The Result of Ablation Study}
	\label{fig:ablation}

\end{figure}
To answer Q3, we conduct an ablation study to verify the effectiveness of our relation discovery module on two multiclassification datasets.
Specifically, we design two ablation mode.
In the first mode, we simply remove the relation discovery module, leaving the task-awareness masking classifier modeling directly processes the node feature. 
Moreover, we examine the performance of different parameter selection proportion $c$ for the task-awareness masking classifier modeling and name the ablation model with $c=0.1$ as selective\_0.1 for example. 
In the second mode, we replace our relation discovery module with other graph neural networks and keep the other setting. 
In the experiment on CitationNet, we choose the graph convolutional networks \cite{DBLP:journals/corr/KipfW16}. 
And we choose graph isomorphism network \cite{GIN} for the OGBN-arxiv dataset.

The result of the ablation study is shown in Figure \ref{fig:ablation}. Our method obviously outperforms the other baselines, confirming the positive effects of disentangling latent relations.
The method in the first mode exhibits a more serious drop than the other methods, which means the general GNN method hardly captures the cross-task information.
Meanwhile, the RAM-CG constantly performs better than the methods in the second mode, showing the proper usage of graph topology is worthy.

\section{Conclusion}

In this work, we propose the Relation-aware Adaptive Model for Continual Graph Learning (RAM-CG).
Our RAM-CG model captures both the task-level relatively stable node relation and task-specific information.
A relation analysis module implicitly disentangles the possible relation between node pairs and generates high quality node embedding  based on the node relation and graph topology.
Meanwhile, a task-awareness masking classifier extracts the task-specific information with selective parameter activation.
Extensive experiments on the benchmark datasets confirm the design validity and show the superiority of our RAM-CG.

\bibliographystyle{unsrt}
\bibliography{main}
\end{document}